\documentclass[sigconf,natbib=true,anonymous=false]{acmart}

\usepackage{multicol}
\usepackage{algorithm2e}
\RestyleAlgo{ruled}
\usepackage{multirow}
\usepackage{tikz}
\usepackage{pgfplots}
\usepackage{capt-of}

\AtBeginDocument{%
  \providecommand\BibTeX{{%
    \normalfont B\kern-0.5em{\scshape i\kern-0.25em b}\kern-0.8em\TeX}}}
\setcopyright{acmlicensed}
\copyrightyear{2018}
\acmYear{2018}
\acmDOI{XXXXXXX.XXXXXXX}

\acmConference[]{}{}{}
\acmISBN{978-1-4503-XXXX-X/18/06}

\begin{document}

\title{NarrationDep: Narratives on Social Media For Automatic Depression Detection}
\author{Hamad Zogan\(^1\), Imran Razzak\(^2\), Shoaib Jameel\(^3\), Guandong Xu\(^1\)}
\email{hamad.zogan@gmail.com, imran.razzak@unsw.edu.au, M.S.Jameel@southampton.ac.uk, guandong.xu@uts.edu.au}
\affiliation{%
  \institution{University of Technology Sydney\(^1\), University of New South Wales\(^2\), University of Southampton\(^3\)}
  \city{}
  \state{}
  \country{}
}

\begin{abstract}
Social media posts provide valuable insight into the narrative of users and their intentions, including providing an opportunity to automatically model whether a social media user is depressed or not. The challenge lies in faithfully modelling user narratives from their online social media posts, which could potentially be useful in several different applications. We have developed a novel and effective model called \texttt{NarrationDep}, which focuses on detecting narratives associated with depression. By analyzing a user's tweets, \texttt{NarrationDep} accurately identifies crucial narratives. \texttt{NarrationDep} is a deep learning framework that jointly models individual user tweet representations and clusters of users' tweets. As a result, \texttt{NarrationDep} is characterized by a novel two-layer deep learning model: the first layer models using social media text posts, and the second layer learns semantic representations of tweets associated with a cluster. To faithfully model these cluster representations, the second layer incorporates a novel component that hierarchically learns from users' posts. The results demonstrate that our framework outperforms other comparative models including recently developed models on a variety of datasets.
\end{abstract}


\maketitle

\section{Introduction}
\label{sec:ch7-intro_background}
The narrative is not the story itself but a representation or a particular form of the story. The narrative transforms the story into knowledge or information, and each event is a unit of knowledge that provides insight into the story. Specifically, narratives can be presented through a sequence of written or spoken words \cite{wang2017predicting}. A narrative connects events, showing their patterns, and relating them to each other or specific ideas, themes, or concepts.

The social media narrative is a long story broken up into posts for social media outlets such as Facebook, Twitter, LinkedIn, and Instagram \cite{dayter2015small}. While the individual posts are short, together, they can craft a longer story with a theme. The narrative plays an important role in the understanding of events that occur in news or stories \cite{chambers2008unsupervised} or in our daily lives and contains many social as well as moral norms.

Narrative in social media refers to how individuals, organizations, and other entities use social media platforms to tell stories and convey information \cite{lund2018power, page2015narrative}. This can include text, images, videos, and other forms of content and can be used for a wide range of purposes, such as building brand awareness, promoting products or services, and engaging with audiences.

\begin{figure}
    \centering
    \includegraphics[scale=0.13]{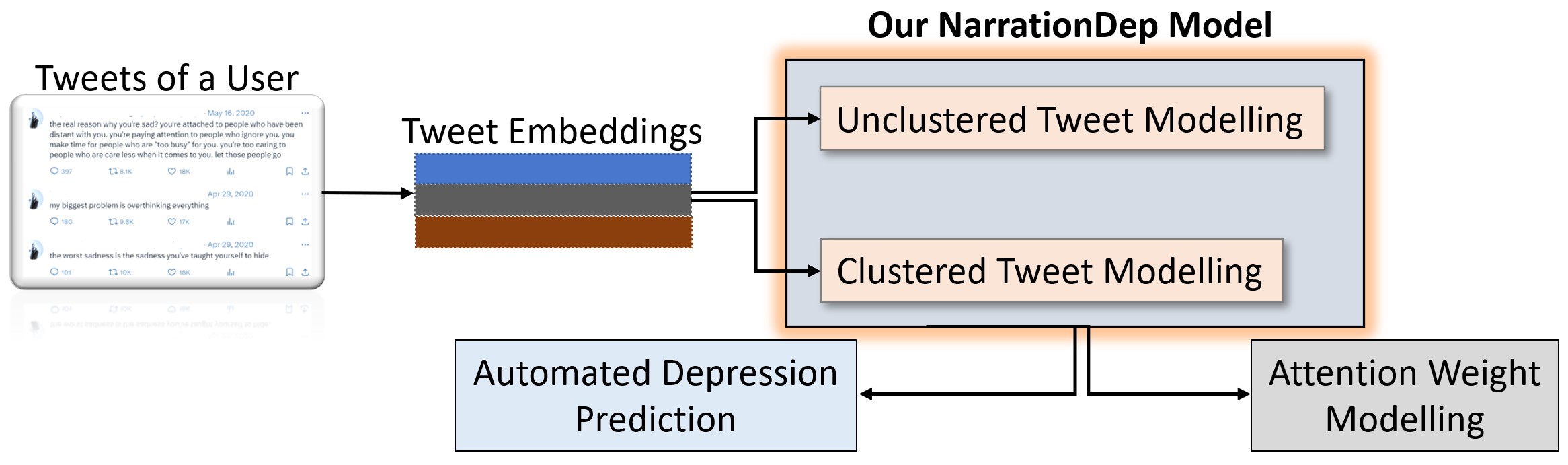}
    \caption{\texttt{NarrationDep} models user narratives to automatically detect depression-related narratives in user-generated text. By modelling attention weights, \texttt{NarrationDep} provides qualitative insights into the state of mind of potentially depressed users. It takes user tweets as input and produces a prediction of whether the user demonstrates signs of depression. Additionally, the learned attention weights reveal key elements captured by the model, aiding in interpretability. Notably, \texttt{NarrationDep} not only models user tweets as a whole but also considers different clusters of tweets and their corresponding cluster assignments.}
    \label{toy_2}
\end{figure}

One important aspect of narratives in social media is the role they play in shaping people's perceptions and understanding of events and issues \cite{holmstrom2015narrative}. Social media narratives can be used to inform, entertain, persuade, and mobilize people around particular issues or causes \cite{seo2018narrative}. They can also be used to shape public opinion, influence political discourse, and even shape public policy \cite{barassi2020social,hasib2023depression}. The ability to generate a narrative can enhance the user's experience and make it more engaging and interactive. Of particular value to organizations such as governments (for example, for defence purposes) is understanding user narratives \cite{ramage2019narrative} on social media, which is one of the most captivating approaches to understanding the users' intentions. This is of interest to several organizations, such as defence agencies, advertisement agencies, and healthcare providers.

Therefore, our goal is to study to what extent it is possible to extract features associated with modelling narratives, including their explanations \cite{ghosh2023attention}, using social media posts. These narratives will vary significantly for people who are going through depression. While challenging, it is indeed possible to reliably model depressed users' narratives using their publicly available posts on social media websites.

Detecting depression through analysis of a user's language on social media platforms is a complex task \cite{zogan2021depressionnet, nguyen2022improving, zhang2022psychiatric, lan2024depression,} and has received attention. One approach to discerning depressive language on more general social media channels, like Twitter, involves identifying statements of diagnosis such as \textit{``(I'm/I was/ I am/ I've been) diagnosed with depression''} \cite{coppersmith2015adhd, shen2017depression, wang2024explainable}. However, this method has limitations since individuals who openly publish a diagnosis on Twitter may not be attempting to conceal their condition and may not represent the group of undiagnosed individuals that this approach seeks to identify \cite{zogan2022explainable}.

Detecting mental health status from social media data may be easier if individuals tweet about disorder-related issues \cite{garg2023mental}. For example, \citet{zogan2021depressionnet} presented a deep network that summarizes a user's relevant post history to identify the most crucial user-generated tweet automatically for a depressed user. This framework allows for greater focus on the most relevant information during model training. Despite these advances, previous studies have faced challenges due to the nature of social media data, which is often unconnected and unordered.

To better understand the content of depressed individuals and differentiate them from non-depressed individuals, it is critical to extract a chain of events that tells a story from a particular point, the topic of these events, and their associated sentiments. This requires an understanding of a user's posts, including narratives such as plot, character, conflict, theme, and tone \cite{herman2009basic}. In the context of social media, however, it is sufficient to comprehend a few of these elements, such as the plot, which refers to the series of events that make up the story in a user's tweets, and the tone or mood conveyed by a user's tweets.

It is common on social media resources that several user's posts converge to a story. However, the events in the story are not sequential or coherent, which makes it challenging to understand as if it resonates like a regular story. To analyze a user's post story, we need to understand their narrative and model the narrative elements of their posts such as the plot mentioned above. People often use social media to engage in events and activities that they find interesting and meaningful. Some posts about specific events may appear in a chronological feed, which displays content in a timeline format. However, events may be shared in different chronologies, creating a social media narrative that tells a long story through various postings on platforms such as Twitter and Facebook. Despite their brevity, these social media posts can be combined to create a broader story with a theme.

Our study explores using social media posts to extract features that can provide a narrative explanation of a series of events and how these features may differ when comparing individuals with a mental disorder. However, analyzing a vast number of user tweets poses a challenge as we need to identify hidden patterns within them. To address this, we can employ clustering analysis, which partitions data into coherent groups, revealing patterns and groupings within our data. The final output would consist of clusters, each containing a set of tweets. Our objective is to determine to what extent social media posts can reveal such features to understand and generate a narrative of a series of events.

Although AI and NLP systems have yet to achieve the goal of learning and generating narratives for regular text, it would be a significant challenge to achieve this aim for social media data, such as posts from depressed users on Twitter. It requires an understanding of crucial narrative elements, how they evolve and extract a story, and identifying relevant tweets amidst the variety of tweets. Nevertheless, narrative tweets can help identify patterns and trends in the way individuals talk about their mental health, as we have shown in (Section \ref{Case_study}), aiding in understanding the experience of depression. Additionally, it can be a valuable tool for self-expression and self-reflection, which can benefit individuals struggling with depression. However, it is important to keep in mind that tweets do not substitute professional diagnosis or treatment. Extracting narrative tweets can be a way to identify individuals at risk of depression and reach out to them for help.

We have developed a new framework called \texttt{NarrationDep}, whose high-level design is depicted in Algorithm~1 and Figure~\ref{toy_2} to model user narratives and automatically depressed users online \cite{yadav2023review}. Our framework is designed to identify Twitter users who show signs of depression and even model explanations from their posts. \texttt{NarrationDep} comprises two components: One is for representing user content (tweet history) using two types of attention (word and tweet level). Another for clustering tweets using a novel hierarchical attention network called the Hierarchical Attention-based Clustering Network (HACN).

We have experimented with our model by using user tweets and semantic clusters individually and found that integrating both components as a unified model improves performance. Our framework is designed for narrative-based depression detection, with the ability to extract narrative tweets and learn explainable \cite{han2022hierarchical} information from a user's content. \texttt{NarrationDep} returns the user's cluster tweets with attention scores, identifying the most significant clusters that contributed to the classification of depression. Each tweet in those clusters represents a narrative tweet in the user's content.

Our research has made significant contributions to the field of automatic depression detection on social media, including:

\begin{enumerate}
    \item We are the first to study the problem of automatically modelling narratives on social media for depression. This is a novel and important research area, as it has the potential to enable early detection and intervention for people at risk of depression.
    \item We have developed a novel hierarchical framework that models social media posts via clustering as depicted in Figure~\ref*{cluster}. This framework takes into account the hierarchical structure of social media data, where words form tweets, tweets from user profiles, and user profiles form narratives. This allows us to capture more complex and nuanced relationships between the different elements of social media data, which is essential for accurate depression detection.
    \item We develop a principled method to jointly exploit user content and cluster tweets to capture an explainable cluster for user narrative understanding. This is important because it allows us to understand why our model makes the predictions it does, which is crucial for developing trustworthy and reliable depression detection systems.
    \item We demonstrate that our method outperforms recently developed techniques based on automatic text summarization. This shows that our approach is more effective at capturing the complex and nuanced relationships between the different elements of social media data, which is essential for accurate depression detection.
\end{enumerate}

\section{Related Work} \label{Related_Work}
In this section, we summarize closely related work and highlight the novelty of our model compared to the previous works.

\subsection {Text mining and Narrative} 
Text mining has been widely used in mental health research, especially in depression studies \cite{wang2016text, wu2020using, payton2018text}. Researchers have used text mining to analyze a variety of text data sources, including social media posts, electronic medical records, and other online content, to identify patterns and trends in the way people talk about depression \cite{mossburger2021exploring, wu2020using, geraci2017applying, lu2020treatment, bhattacharjee2018depression}.

Text mining has been used to identify markers of depression, such as specific words or phrases \cite{eichstaedt2018facebook}, as well as to understand how people talk about depression in different contexts, such as across different demographic groups or in different social media platforms \cite{kleppang2018association, klosterman2014text, zhou2021detecting}.

Overall, text mining has been a powerful tool for gaining new insights into the way that people experience and talk about depression. This knowledge can help mental health professionals better understand the condition and develop more effective treatment strategies.

Text mining and narrative are closely related, as text mining techniques are often used to analyze and understand narrative texts. Narrative text mining specifically applies text mining techniques to extract structured information from narrative texts, such as identifying key elements of the narrative (e.g., characters, events, and themes) and understanding the relationships between these elements \cite{goh2017construction, gupta2009survey}. By using text mining techniques to analyze narrative texts, researchers and analysts can gain insights into the text and its meaning that would be difficult to obtain through manual analysis. Additionally, text mining can also be used to extract narrative summaries \cite{srivastava2016inferring, ghodratnama2021intelligent} and classify narratives \cite{goh2017construction, zhang2005narrative}.

In a recent study, \cite{catipon2021different} studied the distinctions between conservative narratives on Twitter and found that media bias levels were more pronounced on Parler than on Twitter. They also noticed that subjects that are more controlled on Twitter had vastly different perspectives on Parler and that well-known news sources were more politically diverse on Parler. Additionally, \cite{hussain2021stories} created a narrative visualization tool to assist analysts in identifying diverse themes and related narratives being discussed on various blogs. In \cite{rosa2021theaitre} the authors introduced THEaiTRE 1.0, a system based on GPT-2 for generating theatre play scripts. However, extracting a narrative from social media, such as Twitter, can be challenging because the text is often unstructured, with a limited number of characters and a high degree of noise and variation in language, making it difficult to extract meaningful information. Narrative tweets, or tweets that tell a story or describe an event or experience, can potentially be a valuable tool for understanding depression. Therefore, in this paper, we will present, for the first time, extracting narrative tweets and using them to understand depression.

\subsection{Depression Detection on Social Media}
Social media platforms have the potential to assist in identifying and developing methods for diagnosing major depressive disorders \cite{rashida2023social, santos2023setembrobr}. Researchers have been examining the impact of social media on depression prediction as these platforms offer the ability to evaluate an individual's state of mind, thoughts, and content \cite{cai2023depression, adarsh2023fair}. Studies have demonstrated that analyzing user content and textual information on social media can be effective in detecting depression and other mental illnesses \cite{coppersmith2014quantifying, de2013predicting, de2014characterizing}. 

Wang et al., \cite{wang2017detecting} proposed a method for automatically gathering people who described themselves as having an eating disorder in their Twitter profile descriptions to detect eating disorders within social media communities. The authors collected features of linguistics from users for psychometric qualities, and they used similar settings described in \cite{ramirez2018early, kumar2019anxious, soton423226}. From Twitter and Weibo, the authors collected 70 features. They took these characteristics from a user's profile and user engagement characteristics like many followees and followers. Zogan et al., \cite{zogan2022explainable} recently introduced a new model for detecting depressed users using social media, which is an interplay between multilayer perceptron (MLP) and hierarchical attention network (HAN). MLP was used to encode users' online behaviour, while HAN encoded all user tweets at two levels: word-level and tweet-level. They determined each tweet and word weight and extracted characteristics derived from user tweets' semantic sequences. In another work by \cite{zogan2021depressionnet}, the authors argued that using all user tweets to identify a depressed user is ineffective and could even degrade a model performance; therefore, they proposed a new summarization framework interplay between extractive and abstractive summarization to generate a shorter representation of user historical tweets and help to reduce the influence of content that may not eventually benefit the classifier.

Recently, large language models have been used to automatically detect depression \cite{wang2024explainable, xu2024mental, hua2024large, farruque2024deep, xia2024depression, tao2024depmstat}. While these models are effective, they directly do not model user narratives. Our novel method combines grouping tweets with a user's previous posts to enhance depression identification. Additionally, our strategy of selecting relevant content by using a clustering method enables our model to concentrate solely on the most vital information and grasp the narrative element of a depressed user.

\SetKwComment{Comment}{/* }{ */}
\begin{algorithm}[t]
\caption{Our \texttt{NarrationDep} model for a user.}\label{alg:narrationdep}
1: \KwData{User $U_i \subset D$}
2: \KwData{$U_i \equiv \{u_1,u_2,\cdots,u_L\} $}
3: \KwResult{User depression label \(\hat{y}\) \{\textit{Depressed} or \textit{Non-Depressed}\}}
4: \(\vec{U}_i \equiv \{\vec{u}_1,\vec{u}_2,\cdots,\vec{u}_L\}\) \(\leftarrow\) Semantic representations: SBERT(\(U_i\))\;
5: \(c\) \(\gets\) Cluster \(\vec{U}_i\) into \(k\) semantic units\;
6: \(\vec{\gamma}\) \(\leftarrow\) Hierarchical Attention Network: HAN(\(\vec{U}_i\)) for unclustered tweets \;
7: \(\vec{\beta}\) \(\leftarrow\) Hierarchical Attention Network of Tweet Cluster: HCAN(\(c\))\;
8: Jointly learn \(\vec{\theta} \leftarrow\) \(\vec{\gamma}\) \(\bigoplus\) \(\vec{\beta}\) the tweet and event/cluster co-attention features\;
9: Learn a neural classifier \(\hat{y} \leftarrow F(\vec{\theta}, y)\) \;
10: Output: \(\hat{y}\) \(\rightarrow\) User depressed or not\;
\end{algorithm}\label{alg:MYALG}

\section{Our Novel NarrationDep Model}\label{explainable}
\subsection{Notations}
Our \texttt{NarrationDep} model comprises different components, which we have depicted in Algorithm~1. We consider a dataset $D =[1,2,...,U]$ made of $U$ users, and that a user has \(L\) tweets. The clustering method groups the user's tweets into \(E\) clusters $\{c_i\}^{E}_{i=1}$, and each cluster $c_i$ contains $J$ tweets $\{t_{ij}\}^{J}_{j=1}$. Each cluster is, thus, represented by its tweets and each tweet is represented by the sequence of \(d\)-dimensional embeddings of their words $c_i=\{w_{i11},w_{i12},...,w_{ijm} \}$, with $m \in [1,M]$ and $j \in [1,J]$, respectfully, represents the number of words in a tweet and the number of tweets in the \(i^{\text{th}}\) cluster. Suppose the input which resembles all user tweet $U$ be represented as $\{b_n\}^{L}_{n=1}$ where $L$ is the total number of a user tweets and each tweet $b_n = \{w_{n1},w_{n2},...,w_{nq} \}$ with $q \in [1,Q]$ represent, the maximum number of words in a user tweet.

\subsection{Model Design Description}
\begin{figure}
  \centering
  \includegraphics[scale=0.3]{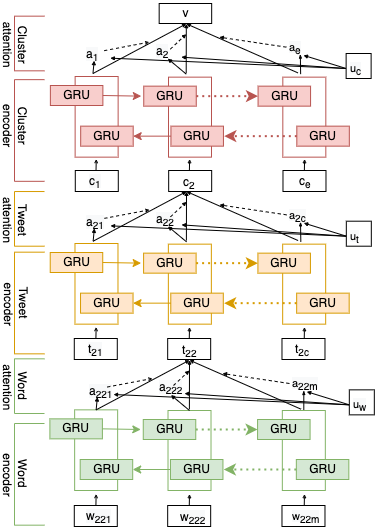}
  \caption{An illustration of our novel Hierarchical Attention Network for Clustering of Tweets (HACN) model.}
  \label{cluster}
  \end{figure}

In \texttt{NarrationDep}, each user's tweet history is pre-processed first. We obtain the semantic representations/features of every tweet using the popular SBERT-base \cite{reimers2019sentence} model as depicted in Line 3. SBERT helps us obtain a reliable semantic representation of every user-generated content. We cluster these semantic representations using a clustering algorithm such as \(k\)-means or HDSBCAN. The reason to cluster is to group similar instances and reduce redundancies which helps make the model reliable. The original set of tweets and the clustered tweets with their cluster assignments are then input into two different components. One component is the Hierarchical Attention Network of User Tweets (HAN) model that jointly models the word and the tweet encodings for every user tweet. The other component is the Hierarchical Attention Network for Clustering of Tweets (HCAN) also depicted in Figure~\ref{cluster}. This component takes as input the user tweet representations along with their cluster assignments. In addition to what HAN learns, HACN also learns the tweet encodings from its inputs. The novel HACN component models clustered tweets to not only gain insights at the tweet level but also at the cluster level. As previously mentioned, tweets comprise words, and each user has a set of tweets that may convey a story or narrative. Clustering these tweets can provide insights into different user narratives, such as those related to depression. The outputs from HAN and HACN are then fused to obtain a joint co-attention representation of users and the associated events that are then input to the depression classifier that outputs the probability of the user being depressed or not.

While similar ideas to jointly model vector representations have been explored in previous works such as \cite{jameel2015unified}, our method is substantially different since we use short text descriptions from Twitter. We have developed a co-attention framework that models the importance of the instances fed as features to the neural classifier that takes inputs from local tweets and clustering of user tweets. This unified approach enables \texttt{NarrationDep} to outperform existing methods by mitigating error propagation from one stage to another, which is common in cascaded methods \cite{jameel2015unified}.

The co-attention layer not only aids in reliable prediction but also plays a crucial role in narrative interpretability. Since a user's tweets may be similar due to the homogeneity of their interests, we adopt a clustering approach to cluster tweets into different meaningful clusters. Clustering also helps us model the narratives associated with the user and mitigate redundancy issues \cite{zogan2021depressionnet}.

To learn context-sensitive information from the annotations, \texttt{NarrationDep} uses a bidirectional Gated Recurrent Unit (BiGRU) as the word-level encoder. GRU is a type of Recurrent Neural Network (RNN) that can capture sequential information and the long-term dependencies of sentences. The BiGRU model, which uses two gate functions (reset and update gates), extracts contextual features that consider both preceding and following words in sequential textual data. It consists of a forward $\overrightarrow{GRU}$ and a backward $\overleftarrow{GRU}$ that process forward and backward data, respectively. The annotation $w_{ijm}$ represents the word $m$ in a tweet $j$ and cluster $i$, which contains \(M\) words. Each word of a user post (tweet) converts to a word embedding $x_{ijm}$.

There are several options for obtaining word embeddings, such as learning embeddings within the model by first one-hot encoding or exploiting static word embeddings like GloVe \cite{pennington2014glove}. Nowadays, it is also common to obtain contextualized word embeddings from models such as BERT \cite{devlin2018bert} and others. Since we are already using the SBERT model in this framework, we initialized the word embeddings by obtaining word representations from the SBERT model. Besides, there are several key advantages of using contextualised embeddings in our framework rather than static word embeddings \cite{devlin2018bert, neumann2018deep}.

The formulation for the GRU model is as follows:

\begin{equation}
  {\overrightarrow{ h_{ijm}^w}} = \overrightarrow{GRU}\left( {{x_{ijm}}} ,\overrightarrow{ h_{ij(m-1)}}\right),  m \in \{1,..., M\}
\end{equation}

\begin{equation}
  {\overleftarrow{ h_{ijm}^w}} = \overleftarrow{GRU}\left( {{x_{ijm}}} ,\overleftarrow{ h_{ij(m-1)}}\right),  m \in \{M,..., 1\}
\end{equation}

The combination of the hidden state that is obtained from the forward GRU and the backward GRU \(\overrightarrow {h_{ijm}^w} \) and \( \overleftarrow {h_{ijm}^w}\) is represented as \( h_{ijm}^w \) $=\left[{\overrightarrow{ h_{ijm}^w}} \oplus {\overleftarrow{ h_{ijm}^w}}\right]$, which carries the complete tweet information centred around ${x_{ijm}}$. We explain the attention mechanism which involves introducing a trainable vector $u_{ijm}$ for all words to capture global word representations. The annotations \({h_{ijm}^w}\) form the foundation for attention which begins with another hidden layer. The model learns by first randomizing biases ($b_w$) and weights ($W_w$) through training. The vector $u_{ijm}$ for the annotations is represented as follows:

\begin{equation}
 u_{ijm} = tanh (W_w h_{itm}^w + b_w)
\end{equation}

The product ${u_{ijm} u_w}$ (${u_w}$ is randomly initialized) is expected to signal the importance of the $m$ word and normalized to an importance weight per word $\alpha_{ijm}$ by a softmax function:

\begin{equation}
\alpha_{ijm}= \frac{\exp(u_{ijm} u_w)}{\sum\limits_{m}\exp(u_{ijm} u_w)}
\end{equation}

A weighted sum of word representations concatenated with the annotations previously determined called the tweet vector $t_{ij}$, where $\alpha_{ijm}$ indicating importance weight per word: 

\begin{equation}
t_{ij}= \sum\nolimits_{m} \alpha_{ijm} h_{ijm}^w
\end{equation}

We now describe the key encoding components used in both HAN and HACN components. There is a cluster encoding component in our model that comprises modelling both tweet and word encoding simultaneously because clusters comprise tweets and tweets comprise words.

\emph{Tweet or post encoding:} To learn the tweet representations $h_{ij}^t$ from a pre-trained tweet vector $t_{ij}$, we capture the information of context at the tweet level. Similar to the word encoder component, the tweet encoder employs the same BiGRU architecture. Hence the combination of the hidden state that is obtained from the forward GRU and the backward GRU \(\overrightarrow {h_{ij}^t}\) and \(\overleftarrow {h_{ij}^t}\) is represented as \(h_{ij}^t \) $=\left[{\overrightarrow{ h_{ij}^t}} \oplus {\overleftarrow{ h_{ij}^t}}\right]$ which captures the coherence of a tweet concerning its neighbouring tweets in both directions. To measure the significance of the tweets, we once more employ the attention mechanism and introduce a context vector at the tweet level $u_t$. The results are computed as follows:

\begin{multicols}{3}
  \begin{equation}
    u_{ij} = tanh (W_t h_{ij}^t + b_t) \nonumber
  \end{equation}\break
  \begin{equation}
    \alpha_{ij}= \frac{\exp(u_{ij} u_t)}{\sum\limits_{j}\exp(u_{ij} u_t)}  \nonumber
  \end{equation}\break
  \begin{equation}
    c_{i}= \sum\nolimits_{j} \alpha_{ij} h_{ij}^t  \nonumber
  \end{equation}
\end{multicols}

\noindent where $c_i$ is the vector that summarizes all the information of tweets in cluster $i$, which encode the word context and sentence context, respectively, and are learned jointly with the rest of the parameters.

Our objective is to cluster tweets that are semantically similar to each other so that we can predict local themes at the cluster level. Clustering the tweets helps us to generate a more coherent and organized representation of them. During the clustering step, we use a technique that groups tweets together based on their semantic similarity. We have chosen the HDSBCAN algorithm for clustering user-generated content \cite{malzer2020hybrid}. This algorithm is suitable for our clustering needs because it clusters tweets hierarchically, uses stability methods to extract flat clusters from the tree, and is beneficial in clustering data of variable densities. We determine the optimal values for the method's parameters through a Bayesian search, which is based on a user-defined objective function and constraints.

Our goal after clustering the user tweets is to model these clusters to obtain cluster encoding, which plays a crucial role in predictive performance and reliable explanations. We have developed a new model, the Hierarchical Attention Based Clustering Network (HACN), inspired by \cite{yang2016hierarchical}, to learn user tweets' representation at the cluster level. Figure \ref{cluster} depicts this new model. We developed this hierarchical framework because words and tweets are context-dependent, meaning that the same words can have different meanings under different contexts. As \cite{bahdanau2014neural} developed, our model includes sensitivity to context information and introduces different levels of attention models. We aim to locate tweets from users that can shed light on the reason for their depression and help detect depression, as they provide explainability. Not all tweets from a user may have the same significance in determining if the user is depressed or not. Therefore, we use an attention mechanism to determine the relevance of tweets about depression and assign attention weights to them, resulting in more accurate and explainable predictions.

We model the tweets in the vector $t$ by implementing a tweet-level attention layer. The product ${u_{i} u_s}$ indicates the significance of the $i$ tweet and is normalized to a per-tweet importance weight $\alpha_{i}$. $s_i$ is a vector that summarizes all the information in a user's tweets.

\begin{equation}
s_{i}= \sum\nolimits_{t} \alpha_{i} h_{i}^t
\end{equation}

\noindent and we summarize our model using the equations below:

\begin{multicols}{2}
  \begin{equation}
    {\overrightarrow{ h_i}} = \overrightarrow{GRU}\left( {{v_i}} ,\overrightarrow{ h_{i-1}}\right)
  \end{equation}\break
  \begin{equation}
    {\overleftarrow{ v_i}} = \overleftarrow{GRU}\left( {{u_i}} ,\overleftarrow{ h_{i-1}}\right)
  \end{equation}
\end{multicols}

\begin{multicols}{2}
  \begin{equation}
    h_{i}=\left[{\overrightarrow{h}{_i}} \oplus {\overleftarrow{h}{_i}}\right]
  \end{equation}\break
  \begin{equation}
    {u_i} = tanh (W_s h_i + b_s)
  \end{equation}
\end{multicols}

\begin{multicols}{2}
  \begin{equation}
    \alpha_{i}= \frac{\exp(u_i u_s)}{\sum\limits_{t}\exp(u_i u_s)}
  \end{equation}\break
  \begin{equation}
    \bar{s_{i}}= \sum\nolimits_{i} a_{i} h_{i}
  \end{equation}
\end{multicols}

\emph{Word encoding}: Similar to the tweet encoder in the cluster tweets encoding above, we represent each word as a fixed-size vector from pre-trained word embeddings. Each word of a user post (tweet) will be converted to a word embedding denoted as $x_{nq}$. The complete tweet information centred around ${x_{nq}}$ is represented as a combination of the hidden state obtained from the forward GRU and backward GRU, $\overrightarrow{h_{nq}^w}$ and $\overleftarrow{h_{nq}^w}$ respectively, represented as $h_{nq}^w = [\overrightarrow{h_{nq}^w} \oplus \overleftarrow{h_{nq}^w}]$. We describe the attention mechanism, which computes a weighted sum of word representations concatenated, called the tweet vector $b_n$, where $\alpha_{nq}$ determines the importance weight per word.

\begin{equation}
b_{n}= \sum\nolimits_{q} \alpha_{nq} h_{nq}^w
\end{equation}

According to previous studies \cite{tsugawa2015recognizing}, it is important to observe people for some time when studying depression. Therefore, when examining a user's tweets within a month of an anchor tweet, it is essential to explore the hierarchical order of the tweets. These sequential tweets may contain valuable semantic information and could potentially boost depression detection when concatenated with a cluster of tweets. User posts contain language cues that differ at the word and tweet level, hence the hierarchical attention network \cite{yang2016hierarchical} may be more accurate in learning user tweet representation for this component, as shown in Figure \ref{cluster}.

To learn tweet representations $h_{n}^b$ from a learned tweet vector $b_{n}$, we capture context information at the tweet level. The tweet encoder employs the same BiGRU architecture as the word encoder component. Therefore, the combination of the hidden state obtained from the forward GRU and the backward GRU, \(\overrightarrow {h_{n}^b} \) and \( \overleftarrow {h_{n}^b}\), is represented as \( h_{n}^b \) $=\left[{\overrightarrow{ h_{n}^b}} \oplus {\overleftarrow{ h_{n}^b}}\right]$, which captures the coherence of a tweet concerning its neighbouring tweets in both directions. To measure the significance of the tweets, we once again employ attention, where a weighted sum of word representations concatenated is called the tweet vector $\hat{b}_{n}$. $\alpha_{n}$ indicates an importance weight per word.

\begin{multicols}{2}
  \begin{equation}
    \hat{b}_{n}= \sum\nolimits_{n} \alpha_{n} h_{n}^w
  \end{equation}\break
  \begin{equation}
    p_{i}=f \bigg(b + \sum\limits_{i=1}^{M} W_i m_i \bigg)
  \end{equation}
\end{multicols}

\noindent where \(f\) represents a nonlinear function, and the output of this function, $p_i$, is a high-level representation that captures the behavioural semantic information. This representation is crucial in the diagnosis of depression. \(m_i\) represents the words in the tweet. The rest of the notations are mentioned in Section~\ref{explainable}.

\subsection{Prediction and Attention Modelling}
\emph{Modelling predictions:} It is necessary to determine whether the user is suffering from depression or not. We have previously described how we process user behaviour features (${p}$) and how we analyze user tweets by modelling the hierarchical structure from the word level and tweet level (${s}$). Subsequently, we combine both components to create a feature matrix of user behaviour features and user tweets to use in our classification task as follows:

\begin{multicols}{2}
  \begin{equation}
    {p = p_1,p_2, .... , p_M} \in {\mathbb{R}^{1d \times M}}
  \end{equation}\break
  \begin{equation}
    s = s_1,s_2, .... , s_n \in {\mathbb{R}^{2d \times n}}
  \end{equation}
\end{multicols}

\noindent We further unify these components \(p\) and \(s\) together as demonstrated in the equation below, which is denoted as $[{p},{s}]$. The output of such a network is typically fed to a sigmoid layer for classification as follows:

\begin{equation}
\hat{y}= sigmoid (b_f+[{p},{s}]W_f)
\end{equation}

\noindent The predicted probability vector, $\hat{y}$, is composed of $\hat{y_0}$ and $\hat{y_1}$, which represent the predicted probability of the label being 0 (not depressed) and 1 (depressed user) respectively. Our objective is to minimize the cross-entropy error for each user with the true label, $y$:

\[\mathrm{Loss} = - \sum_{i} y_i \cdot \mathrm{log}\; {\hat{y}}_i \]

\noindent where \(\hat{y}_i\) is the predicted probability and \(y_i\) is the ground truth label (either depression or non-depression) user.

Our objective is to also identify a particular narrative from a user's attention function of why they may be experiencing depression. Clarity in the explanation can also aid in the detection of depression. The hierarchical attention mechanism, which we described earlier for cluster encoding, is an effective technique for assigning high weights to specific elements of the narrative. Moreover, the level of the user's topic's attention is learned through attention weights. Since tweets in a cluster are assigned different weights based on the attention map, our model can extract relevant and contextual information from the cluster by modelling the attention weights. In general, the attention map of our model can identify the most significant theme that is related to a depressed user and their corresponding group of tweets. Therefore, clusters with high attention weight are crucial and likely to highlight a user's depression.

\section{Experiments and Results}
We conducted thorough experiments to evaluate the superiority of \texttt{NarrationDep} in comparison to other robust models, utilizing both quantitative and qualitative methodologies.

\subsection{Comparative methods}
In a recent study, Zogan et al., \cite{zogan2021depressionnet} developed a new automatic depression detection framework that utilizes both extractive and abstractive summarization techniques to condense user tweets into a shorter representation. By reducing redundant content, this approach makes the set of tweets more focused on a single theme. The authors compared the effectiveness of this method with other models, including convolutional neural networks with attention (CNN-Att) and Bidirectional Gated Recurrent Neural Networks with Attention (BiGRU-Att), as well as three pre-trained transformer models -- XLNet \cite{yang2019xlnet}, BERT \cite{devlin2018bert}, and RoBERTa \cite{liu2019roberta}. We evaluate \texttt{NarrationDep} against these methods, including the ``DepressionNet'' model developed in \cite{zogan2021depressionnet}. The language models were fine-tuned on the dataset used in the study as adopted in \cite{zogan2021depressionnet}.

We employed multiple neural network architectures to classify user tweets for depression detection. First, we used the BiGRU model with an attention mechanism to obtain tweet representations. Then, we utilized the CNN model with an attention mechanism to represent user tweets and capture the semantics of various convolutional window sizes. Next, we implemented a hierarchical attention neural network architecture to identify depression in user postings. This model encrypts the first user postings by paying attention to both the words in each tweet and the tweets themselves. Finally, we improved the identification accuracy by using hierarchical convolutional networks (HCN) instead of GRU, similar to HAN. While there are other related models such as the Latent Dirichlet Allocation (LDA) model, we note that recent works have shown that language models implicitly encode the latent topic information \cite{talebpour2023topics}.

Given the recent advancements in depression modelling since \citet{zogan2021depressionnet}, we have compared our model with a range of newly developed language modelling-based approaches. These models range from MentalBERT \cite{ji2021mentalbert} to the recently developed technique using LLMs \cite{yang2023towards}.

\begin{table}
  \centering
    \caption{Summary of labelled data used to train depression model} 
    \begin{tabular}{l|c|c}
    \toprule
    \toprule
    Description & Depressed & Non-Depressed \\
    \midrule
    Number of users & 2159  & 2049 \\
    Number of tweets & 447856 & 1349447 \\
    \bottomrule
    \bottomrule
    \end{tabular}%
  \label{tab:data}%
\end{table}%

\subsection{Datasets}
\label{dataset_design}
The effectiveness of various models in detecting depression was evaluated using Shen et al. pre-COVID dataset \cite{shen2017depression}, which is presented in Table \ref{tab:data}. This dataset has been widely used by the research community \cite{zogan2021depressionnet} to automatically detect depression. The dataset contains Twitter posts from users, with each user being labelled as either depressed or non-depressed. The authors identified depressed users based on the content of their posts from 2009 to 2016, resulting in approximately 2000 depressed users and approximately 400,000 tweets. Non-depressed tweets were collected in December 2016, comprising 2000 users and more than a million tweets. Users with less than ten tweets were removed from the dataset during preprocessing to ensure accurate results. The dataset was then randomly split into training and test sets with a ratio of 80:20, using 5-fold cross-validation for evaluation. 

We also compare \texttt{NarrationDep} using other recent depression datasets that are summarised in \cite{ji2021mentalbert}. These datasets are eRisk T1, CLPsych, and Depression\_Reddit (denoted as Dep\_Red in Table~\ref{recent-comparisons}).

\begin{table*}[ht]
\begin{minipage}[b]{0.56\linewidth}
  \centering
  \caption{Comparison of depression detection performances.}
  \scalebox{0.7}
  {
    \begin{tabular}{lp{18.32em}cccc}
    \toprule
    \toprule
    \multicolumn{1}{p{8.635em}}{\textbf{Feature}} & \textbf{Model} & \multicolumn{1}{p{4.045em}}{\textbf{Precision}} & \multicolumn{1}{p{4.045em}}{\textbf{Recall}} & \multicolumn{1}{p{4.045em}}{\textbf{F1-score}} & \multicolumn{1}{p{4.045em}}{\textbf{Accuracy}} \\
    \midrule
     \multicolumn{1}{l}{\multirow{6}[2]{*}{User Behaviours}} & \multicolumn{1}{l}{SVM (\citet{pedregosa2011scikit})} & 0.724 & 0.632 & 0.602 & 0.644 \\
          & \multicolumn{1}{l}{NB (\citet{pedregosa2011scikit})} & 0.724 & 0.623 & 0.588 & 0.636 \\
          & \multicolumn{1}{l}{MDL (\citet{shen2017depression})} & 0.790 & 0.786 & 0.786 & 0.787 \\
          & \multicolumn{1}{l}{GRU (\citet{chung2014empirical})} & 0.743 & 0.705 & 0.699 & 0.714 \\
          & \multicolumn{1}{l}{BiGRU} & 0.787 & 0.788 & 0.760 & 0.750 \\
          & \multicolumn{1}{l}{Stacked BiGRU} & 0.825 & 0.818 & 0.819 & 0.821 \\
    \midrule
    \multicolumn{1}{p{8.635em}}{posts + Image} & GRU + VGG-Net + COMMA (\citet{gui2019cooperative}) & 0.900 & 0.901 & 0.900 & 0.900 \\
    \midrule
    \multicolumn{1}{l}{\multirow{6}[2]{*}{Posts Summarization}} & XLNet (base) (\citet{yang2019xlnet})  & 0.889 & 0.808 & 0.847 & 0.847 \\
          & BERT (base) (\citet{liu2019roberta}) & 0.903 & 0.770 & 0.831 & 0.837 \\
          & RoBERTa (base) (\citet{liu2019roberta}) & \textbf{0.941} & 0.731 & 0.823 & 0.836 \\
          & BiGRU (Att) & 0.861 & 0.843 & 0.835 & 0.837 \\
          & CNN (Att) & 0.836 & 0.829 & 0.824 & 0.824 \\
          & CNN-BiGRU (Att)  & 0.868 & 0.843 & 0.848 & 0.835 \\
    \midrule
    \multicolumn{1}{l}{\multirow{6}[2]{*}{Summarization + User Behaviours}} & CNN + BiGRU & 0.880 & 0.866 & 0.860 & 0.861 \\
          & BiGRU (Att) + BiGRU & 0.896 & 0.885 & 0.880 & 0.881 \\
          & CNN-BiGRU (Att) + BiGRU & 0.900 & 0.892 & 0.887 & 0.887 \\
          & BiGRU (Att) + Stacked BiGRU & 0.906 & 0.901 & 0.898 & 0.898 \\
          & CNN (Att) + Stacked BiGRU & 0.874 & 0.870 & 0.867 & 0.867 \\
          & DepressionNet \cite{zogan2021depressionnet} & 0.909 & 0.904 & 0.912 & 0.901 \\
          \midrule
    \multicolumn{1}{l}{\multirow{1}[2]{*}{User Narratives}} & \textbf{NarrationDep} & 0.919 & \textbf{0.914} & \textbf{0.916} & \textbf{0.921} \\
    \bottomrule
    \bottomrule
    \end{tabular}%
}
  \label{tab:dep_Comparison}%
\end{minipage}\hfill
\begin{minipage}[b]{0.3\linewidth}
\caption{Modelling the number of clusters.}
    \centering
    \begin{tikzpicture}[thick,scale=0.6, every node/.style={transform shape}]
\begin{axis}[xlabel=Number of Clusters, ylabel=F1]
	\addplot+[smooth] coordinates 
		{(1,0.858) (10,0.886) (20, 0.892) (30, 0.916) (40,0.906) (50, 0.899)};
\end{axis}
\end{tikzpicture}
\label{cluster_numbers}
\end{minipage}
\end{table*}

\begin{table}[]
\caption{Narration-Dep compared with other recent transformer-based models including MentalLlama under a zero-shot setting.}
\scalebox{0.8}
{
\begin{tabular}{l|llllll}
\hline
\multicolumn{1}{c|}{\textbf{Model}} & \multicolumn{2}{c}{\textbf{eRisk T1}} & \multicolumn{2}{c}{\textbf{CLPsych}} & \multicolumn{2}{c}{\textbf{Dep\_Red}} \\ \hline
 & \multicolumn{1}{c}{Rec.} & \multicolumn{1}{c}{F1} & \multicolumn{1}{c}{Rec.} & \multicolumn{1}{c}{F1} & \multicolumn{1}{c}{Rec.} & \multicolumn{1}{c}{F1} \\ \cline{2-7} 
BERT & 88.53 & 88.54 & 64.67 & 62.75 & 91.13 & 90.90 \\
RoBERTa & 92.25 & 92.25 & 67.67 & 66.07 & 95.07 & 95.11 \\
BioBERT \cite{lee2020biobert} & 79.16 & 78.86 & 65.67 & 65.5 & 91.13 & 90.98 \\
ClinicalBERT \cite{huang2019clinicalbert} & 76.25 & 75.41 & 65.67 & 65.30 & 89.41 & 89.03 \\
DisorBERT \cite{aragon2023disorbert} & 93.45 & 93.59 & 70.88 & 69.77 & 94.40 & 94.50 \\
MentalBERT \cite{ji2021mentalbert} & 86.27 & 86.20 & 64.67 & 62.63 & 94.58 & 94.62 \\
MentalRoBERTa \cite{ji2021mentalbert} & 93.38 & 93.38 & 70.33 & 69.71 & 94.33 & 94.23 \\
MentalLLama \cite{yang2023towards} & 93.89 & 94.00 & 71.34 & 69.99 & 94.01 & 94.85 \\ \hline
NarrationDep-BERT & \textbf{94.12} & \textbf{94.58} & \textbf{72.55} & \textbf{70.12} & \textbf{94.91} & \textbf{95.03} \\
NarrationDep-RoBERTa & \textbf{94.88} & \textbf{94.92} & \textbf{73.01} & \textbf{71.25} & \textbf{95.12} & \textbf{95.89} \\ \hline
\end{tabular}
}
\label{recent-comparisons}
\end{table}
  
\begin{table}
    \centering
    \caption{Performance comparison \texttt{NarrationDep} against closely related models that do not summarise tweets.}
    \scalebox{0.7}
    {
      \begin{tabular}{p{6.785em}cccc}
      \toprule
      \multicolumn{1}{l}{\textbf{Model} } & \textbf{Precision} & \textbf{Recall} & \textbf{F1-Score} & \textbf{Accuracy} \\
      \midrule
       BiGRU  & 0.766 & 0.762 & 0.786 & 0.864 \\
      CNN   & 0.817 & 0.804 & 0.786 & 0.866 \\
      HAN    & 0.870  & 0.844 & 0.856 & 0.875 \\
      HCN   & 0.853 & 0.852 & 0.852 & 0.882 \\
      \midrule
      \textbf{NarrationDep} & \textbf{0.919} & \textbf{0.914} & \textbf{0.916} & \textbf{0.921} \\
      \bottomrule
      \end{tabular}%
      }
    \label{tab:result1}%
  \end{table}%

\begin{table}[]
\centering
    \caption{Performance of our full model compared with HAN and HACN components when they are used independently. It is evident from the results that incorporating both HAN and HACN jointly helps significantly improve the quantitative performance of the model.}
\begin{tabular}{|l|l|l|}
\hline
\multicolumn{1}{|c|}{\textbf{Model}} & \multicolumn{1}{c|}{\textbf{F1}} & \multicolumn{1}{c|}{\textbf{Accuracy}} \\ \hline
HAN & 0.856 & 0.875 \\ \hline
HACN & 0.869 & 0.878 \\ \hline
NarrationDep & \textbf{0.916} & \textbf{0.921} \\ \hline
\end{tabular}
\label{ablation_study1}
\end{table}

\subsection{Evaluation Metrics and Settings}
To gauge the effectiveness of our depression detection approach, we employed a range of established metrics (accuracy, F1-score, recall, and precision) commonly used in similar studies \cite{shen2017depression, zogan2022explainable, zogan2021depressionnet, zogan2023hierarchical, zhou2021detecting, ji2021mentalbert}. Our experiments were conducted in Python 3.6.3 with TensorFlow 2.1.0. We implemented a dropout rate of 0.5 and used the Adam optimization algorithm for both the Heterogeneous Convolutional Network (HCN). The default learning rate (lr) was set to 0.001. As a clustering model, we have used the HDSBCAN clustering algorithm due to its simplicity and scalability. We have tuned the number of clusters based on the validation dataset.
  
\subsection{Quantitative Results}
We present the quantitative results obtained from various models, with the best results highlighted in bold in Table~\ref{tab:dep_Comparison}. When referring to summarization, we followed the approach developed in \cite{zogan2021depressionnet}, which utilizes an abstractive BERT--BART automatic text summarizer to condense tweets before inputting them into the computational model. Additionally, we model user behaviours as described in \cite{zogan2021depressionnet}, as these features can be effectively extracted from the dataset.

We consider these models strong comparators due to their similar approach to the semantic modelling of tweets. However, our model distinguishes itself by adopting a clustering approach to semantically cluster tweets, while \cite{zogan2021depressionnet} uses a tweet summarization strategy to extract the most critical information. Besides that, we have developed a novel HACN framework that models these tweet clusters effectively. Our results demonstrate that \texttt{NarrationDep} outperforms models that rely on tweet summarization for depression detection. This is because our model specifically models semantic clusters, while the comparative models do not. Additionally, the model in \cite{zogan2021depressionnet} heavily relies on the quality of the automatic text summarization model, which can aggressively condense the text and remove relevant information. CNN\_BiGRU-At outperforms other models in F1-score and recall, while XLNet and RoBERTa achieve the best accuracy and precision among all the compared models. Notably, the BiGRU-Att model exhibits high precision, making it reliable for positive cases of depression. However, its F1-score performance falls short of our model's.

In Table~\ref{recent-comparisons}, we compare \texttt{NarrationDep} with recent transformer-based models using them as the feature extractors viz., BERT and RoBERTa. From the results, we observe that our model outperforms these recent methods. The reason why our model quantitatively outperforms them is that the key component in our model is the narration modelling which is absent in the recently developed methods.
  
Table \ref{tab:result1} presents the results of our study, in which we evaluated the performance of various neural network-based methods for classifying user tweets without employing tweet summarization. This means that the comparative models received the original dataset without any summarization techniques applied. Among these models, HAN and HCN, which are hierarchical text classification models, and closely related to \texttt{NarrationDep}, outperformed other neural network-based methods such as BiGRU and CNN. The results demonstrate that even these strong comparative models are less effective than our proposed model when summarization is not utilized. While summarization carries the risk of losing key details within the user dataset, as mentioned earlier, considering the complete set of tweets can also introduce noise that could negatively impact the model's predictive capabilities. Clustering helps to mitigate issues such as redundant information, and jointly modelling clustering and individual user tweets, as in our model, improves performance by reducing error propagation. Additionally, our model considers both global (clustered user tweets) and local structures (full set of user's tweets) in the user-generated data. Our results also outperform the recently developed \cite{aragon2023disorbert} model.

\subsection{Ablation Study}
To evaluate the effectiveness of our model's components individually and their combined impact, we conducted an ablation study using the dataset of \citet{shen2017depression}. We present these results in Table~\ref{ablation_study1}. We first evaluated the model using only the Hierarchical Attention for Clustering Networks (HACN) for tweet clustering, without the Hierarchical Attention Network (HAN) in \texttt{NarrationDep}. We then evaluated the model using only the HAN in \texttt{NarrationDep}, without the HACN component. When the HACN component is used, it outperforms the HCN component. However, when both components are combined in our \texttt{NarrationDep} model, we observe a significant performance improvement. This suggests that the HACN module plays a more crucial role in the overall model effectiveness. This effectiveness stems from the model's ability to capture both individual user tweets and the semantic information acquired from user tweet clusters. This allows the model to leverage:
\begin{itemize}
    \item Local information: Obtained directly from individual user tweets.
    \item Global narratives: Captured from the semantic clusters of user tweets.
\end{itemize}

These combined insights contribute to a more comprehensive and effective data modelling, as demonstrated by the superior performance of the combined model. As the number of tweets per user can be substantial, users may engage in discussions about diverse narratives within their tweet history. Clustering techniques can effectively group these narratives, with each cluster representing a distinct and coherent narrative thread. While the number of clusters has been automatically tuned in our main results, it remains a crucial parameter within \texttt{NarrationDep}. Our findings indicate that \texttt{NarrationDep} attains optimal performance when employing a maximum of 30 clusters. Figure~\ref{cluster_numbers} visually depicts the model's performance about the number of clusters employed.

\subsection{Clustering Algorithm}
We study the importance of clustering algorithms to investigate the impact of the clustering model choice on our \texttt{NarrationDep} model. Specifically, we explore whether efficient flat clustering algorithms, such as k-means, can achieve comparable results to HDSBCAN. To this end, we evaluated the Latent Dirichlet Allocation (LDA) \cite{blei2003latent} topic model as a semantic clustering approach and the popular k-means model. We tuned the number of clusters using the validation dataset. Our observations indicate that flat clustering models do not significantly impact model performance. For example, the k-means model achieved an F1 score of 0.816, while the LDA model achieved 0.815. We also experimented with the hierarchical version of the LDA model \cite{griffiths2003hierarchical} with a tree depth of 3. The results were comparable to those of the HDSBCAN model, with an F1 score of 0.910 compared to 0.916 in the original model.
  
\subsection{Qualitative Study}\label{Case_study}
\begin{figure}
  \centering
  \includegraphics[width=0.70
  \linewidth]{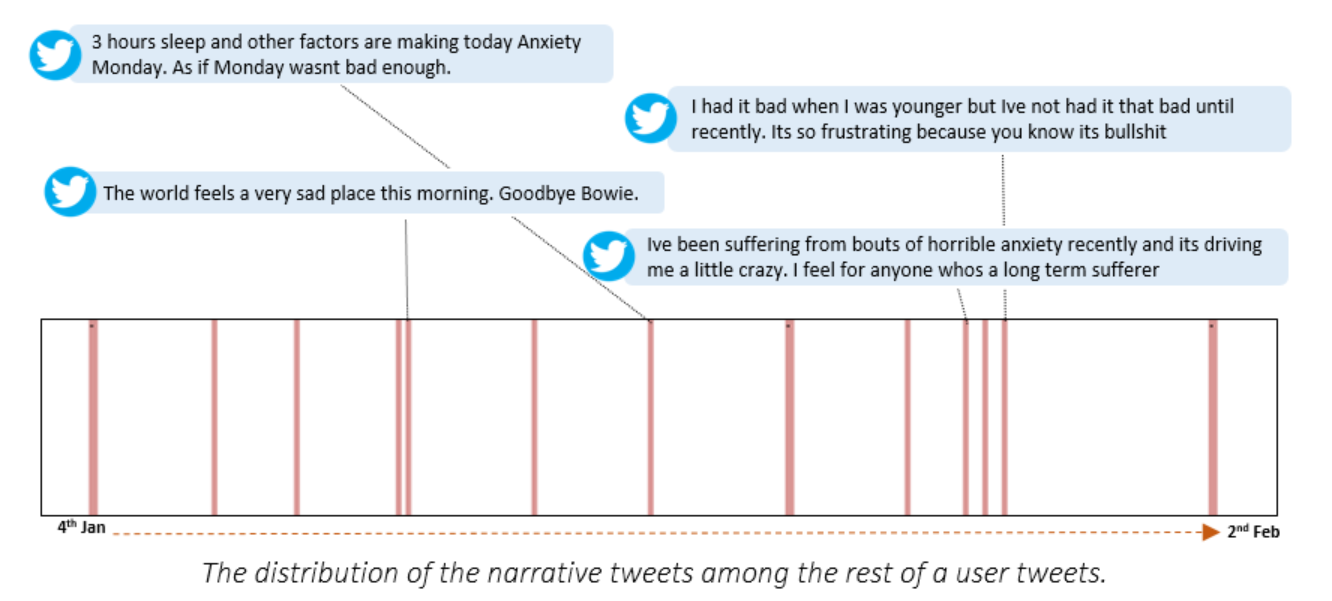}
  \caption{Narrative tweets captured by \texttt{NarationDep}.}
  \label{heatmap}
  \end{figure}

As Algorithm~1 depicts, modelling attention weights allows us to analyze user behaviour and identify depressed mindsets. We qualitatively demonstrate the effectiveness of our \texttt{NarrationDep} model through various analyses that showcase its ability to model depression-related content. To identify tweets contributing to a narrative, we analyze the importance assigned to each tweet by the model (attention weights). Based on our hypothesis, tweets with high attention weights should be central to the narrative. Figure~\ref{heatmap} showcases an attention map for a randomly selected depressed user, highlighting the identified tweets. The figure presents all tweets from the user as a bar chart, where red indicates their contribution to the narrative. Red lines on the chart point to the most relevant cluster for identifying depressed users.

Our qualitative study validates the ability of \texttt{NarrationDep} to extract crucial contextual information about depression from relevant tweet clusters. The attention mechanism effectively identifies tweets that provide key insights into depression, assigning higher weights to those containing depression-oriented language. This enhanced ability to pinpoint key information makes our model highly effective in determining users experiencing depression.

\begin{figure}
    \centering  
    \begin{minipage}[b]{0.30\textwidth}
      \includegraphics[width=\textwidth]{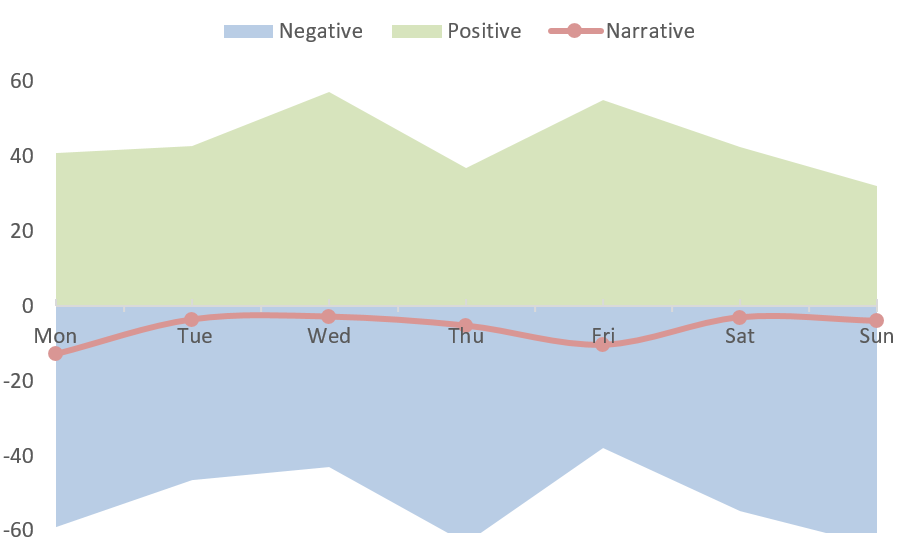}
      \caption{\textit{Analyzing narrative for a user per week. The x-axis represents the days of the week and the y-axis represents the attention weights.}}
      \label{number of tweets}
    \end{minipage}\hfill
      \begin{minipage}[b]{0.30\textwidth}
      \includegraphics[width=\textwidth]{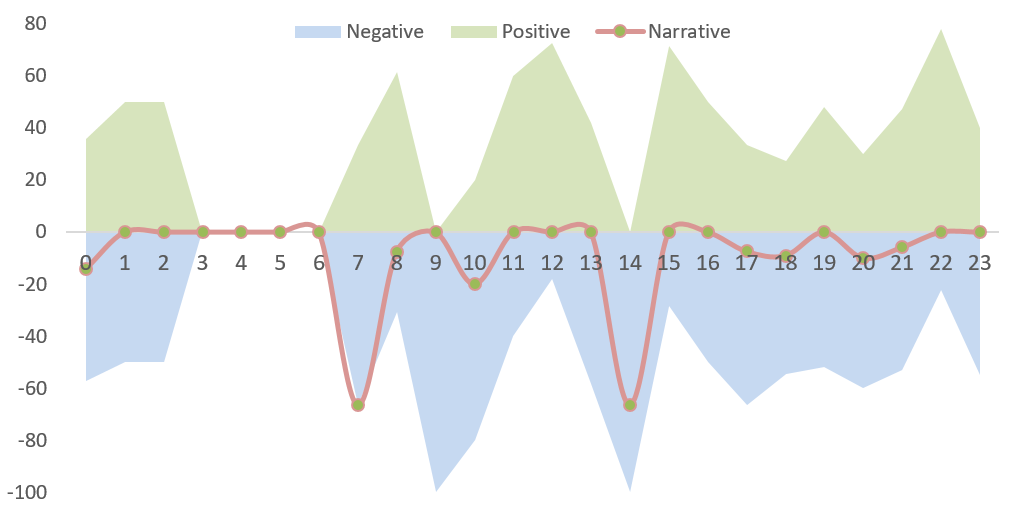}
      \caption{\textit{Analyzing narrative for a user during the 24 hours of the day. The x-axis represents the hours and the y-axis represents the attention weights.}}
      \label{minus}
    \end{minipage}
  \end{figure}

In Figures~\ref{number of tweets} and \ref{minus}, we analyze the temporal dynamics of user narratives, building upon the work of \cite{zhou2022tamfn}. We track how a user's narrative fluctuates between positive and negative sentiments over a week before modelling their daily narrative patterns. Our analysis reveals that depressed users' narratives are predominantly negative, often revolving around themes of depression. On Mondays and Fridays, Figure~\ref{number of tweets} shows that users tend to express more negative sentiments than on other days, a finding that generalizes across several depressed users. Figure~\ref{minus} further reveals that users exhibit peaks in negative sentiment around 7:00 AM and 2:00 PM during the day. Overall, we find that depressed users rarely exhibit a positive state of mind.

We observed that the attention weights extracted from comparative models like DepressionNet, MentalBERT, and SBERT did not offer sufficient insight into the user's mental state. For instance, while \texttt{NarrationDep} predicted words such as \textit{``killing''} with a high likelihood in the phrase \textit{``I feel like [MASK] myself''}, other models predicted the masked token as \textit{``destroying''} using MentalBERT and \textit{``running''} using DepressionNet. Our two-component architecture, which explicitly models the narratives of depressed users, proved more effective in capturing meaningful information. Specifically, unlike DepressionNet, our model avoids significant information loss during the summarization phase.

\section{Conclusion} \label{conclusion}
This paper presents a novel approach for modelling user narratives in social media, specifically focusing on understanding a user's story through their posted content. By leveraging the hierarchical structure of user cluster tweets and employing an attention mechanism, our proposed model, \texttt{NarrationDep}, automatically identifies the most crucial cluster representing a depression-associated narrative and models its underlying interpretability. We achieve this by modelling user input with two key components, utilizing both HAN and HACN to extract diverse features from the same input. Our results demonstrate that \texttt{NarrationDep} with these two attributes significantly outperforms established benchmark models in effectively detecting depressed users.

We will extend NarrationDep into a multi-modal model (building upon work by \cite{yoon2022d}) to incorporate important features from video and image data, which often hold complementary information. To achieve this, we will explore leveraging visual-language models such as DALL-E \cite{ramesh2021zero} or alternative approaches like PaLM or BLIP \cite{driess2023palm, li2023blip}. Additionally, we will develop the model's capability to automatically predict relevant diagnoses for under-studied mental health concerns. Diagnosis will play a key role in the future because the model must not only be reliable but also be explainable \cite{naseem2022hybrid}.

\bibliographystyle{ACM-Reference-Format}
\bibliography{references}
\end{document}